%% file: main.tex
\documentclass[10pt,twocolumn,letterpaper]{article}

\usepackage{cvpr}              

\usepackage[accsupp]{axessibility}  

\usepackage{graphicx}
\usepackage{amsmath}
\usepackage{amssymb}
\usepackage{booktabs}

\usepackage{ulem}
\usepackage{enumitem}

\usepackage[pagebackref,breaklinks,colorlinks]{hyperref}

\usepackage[capitalize]{cleveref}
\crefname{section}{Sec.}{Secs.}
\Crefname{section}{Section}{Sections}
\Crefname{table}{Table}{Tables}
\crefname{table}{Tab.}{Tabs.}

\usepackage{nicefrac}
\usepackage{multirow}
\usepackage{booktabs}
\usepackage{float}
\usepackage{wrapfig}
\usepackage{bbding}
\usepackage{pifont}
\usepackage{color, colortbl}
\usepackage{skak}

\newcommand{\tabincell}[2]{\begin{tabular}{@{}#1@{}}#2\end{tabular}} 
\newcommand{\hr}[1]{\textcolor{black}{#1}}
\definecolor{gray}{gray}{0.9}
\newcommand{\ssymbol}[1]{^{\@fnsymbol{#1}}}

\usepackage{array}
\newcolumntype{A}{>{\centering}p{0.2\textwidth}}
\newcolumntype{I}{>{\centering}p{0.11\textwidth}}
\newcolumntype{B}{>{\centering\arraybackslash}p{0.1\textwidth}}
\newcolumntype{H}{p{0.18\textwidth}}

\def\cvprPaperID{11525} 
\def\confName{CVPR}
\def\confYear{2023}

\newcommand{\hry}[1]{\textcolor{black}{#1}}

\begin{document}

\title{Castling-ViT: \underline{C}ompressing Self-\underline{A}ttention via \underline{S}witching \\\underline{T}owards \underline{Li}near-A\underline{ng}ular Attention at Vision Transformer Inference
\vspace{-1.em}}


\author{Haoran You$^{1,2,\dagger,}$\thanks{Equal contribution. $\dagger$ Work done while interning at Meta Research.}, Yunyang Xiong$^{2,*}$, Xiaoliang Dai$^2$, Bichen Wu$^2$, Peizhao Zhang$^2$, \\Haoqi Fan$^2$, Peter Vajda$^2$, Yingyan (Celine) Lin$^1$\\
$^1$Georgia Institute of Technology, $^2$Meta Research\\
{\tt\small \{\url{haoran.you},\url{celine.lin}\}\url{@gatech.edu}}\\ 
{\tt\small \{\url{yunyang},\url{xiaoliangdai},\url{wbc},\url{stzpz},\url{haoqifan},\url{vajdap}\}\url{@meta.com}}
\vspace{-0.8em}
}
\maketitle

\begin{abstract}
Vision Transformers (ViTs) have shown impressive performance but still require a high computation cost as compared to convolutional neural networks (CNNs), 
one reason is that ViTs' attention measures global similarities and thus has a quadratic complexity with the number of input tokens. 
Existing efficient ViTs adopt local attention or linear attention, which sacrifice ViTs' capabilities of capturing either global or local context.
In this work, we ask an important research question: \textit{Can ViTs learn both global and local context while being more efficient during inference?} 
To this end, we propose a framework called \textbf{Castling-ViT}, which trains ViTs using both linear-angular attention and masked softmax-based quadratic attention, but then switches to having only linear-angular attention during inference.
Our Castling-ViT leverages angular kernels to measure the similarities between queries and keys via spectral angles. And we further simplify it with two techniques: 
(1) a novel linear-angular attention mechanism: we decompose the angular kernels into linear terms and high-order residuals, and only keep the linear terms; and
(2) we adopt two parameterized modules to approximate high-order residuals: a depthwise convolution and an auxiliary masked softmax attention to help learn global and local information, where the masks for softmax attention are regularized to gradually become zeros and thus incur no overhead during inference.
Extensive experiments validate the effectiveness of our Castling-ViT, e.g., achieving up to a \textbf{1.8\%} higher accuracy or \textbf{40\%} MACs reduction on classification and \textbf{1.2} higher mAP on detection under comparable FLOPs, 
as compared to ViTs with vanilla softmax-based attentions.
Project page is available at \href{https://www.haoranyou.com/castling-vit}{here}.
\end{abstract}

\input{sections/1-Introduction}
\input{sections/2-Related_works}
\input{sections/3-Methods}
\input{sections/4-Experiments}
\input{sections/5-Conclusion}

{\small
\bibliographystyle{ieee_fullname}
\bibliography{egbib}
}

\clearpage
\newpage
\input{sections/6-Supple.tex}

\end{document}

%% file: sections/1-Introduction.tex
\vspace{-1em}
\section{Introduction}
\label{sec:intro}
\vspace{-0.3em}

\begin{figure}[t]
    \centering
    \includegraphics[width=\linewidth]{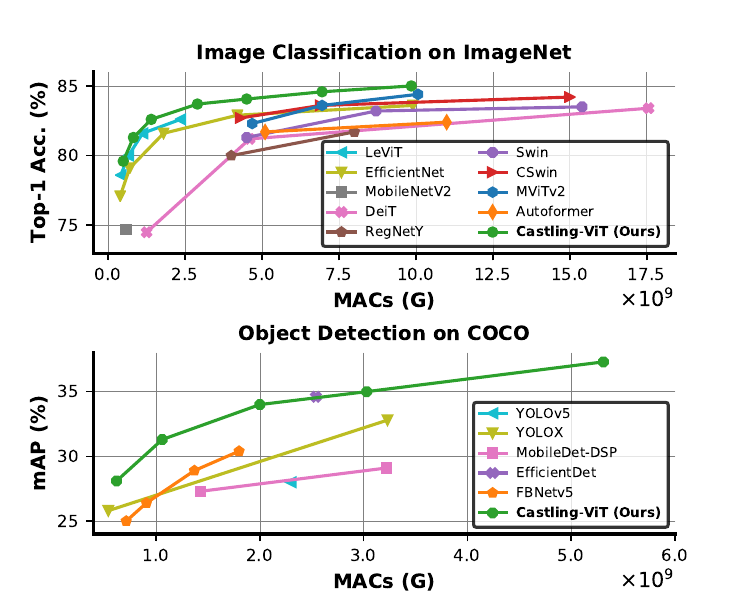}
    \vspace{-2.3em}
        \caption{Castling-ViT over SOTA baselines on (1) ImageNet~\cite{deng2009imagenet} image classification and (2) COCO~\cite{lin2014microsoft} object detection.}
    \label{fig:overall_comp}
    \vspace{-1.8em}
\end{figure}

Vision Transformers (ViTs) have made significant progress in image classification, object detection, and many other applications. It is well recognized that
the superior performance achieved by ViTs is largely attributed to their self-attention modules that can better capture global context \cite{transformer,lite_transformer,vit}.
Nevertheless, ViTs' powerful self-attention module comes at the cost of quadratic complexity with the number of input tokens, causing a major efficiency bottleneck to ViTs' achievable runtime (i.e., inference latency)~\cite{you2022vitcod,xiong2021nystromformer,chen2021scatterbrain,zhu2021long,beltagy2020longformer,kitaev2020reformer,wang2020linformer}.
To mitigate this issue, linear attention designs have been developed to alleviate the vanilla ViT attention’s quadratic complexity.
In particular, existing efforts can be categorized into two clusters: 
(1) ViTs with local attention by restricting the attention window size \cite{liu2021swin,tang2022quadtree}, sharing the attention queries\cite{arar2022learned}, or representing the attention queries/keys with low rank matrices\cite{wang2020linformer}; and
(2) ViTs with kernel-based linear attention, which approximate the non-linearity softmax function by decomposing it into separate kernel embeddings. This enables a change in the matrix computation order for a reduced computational complexity~\cite{katharopoulos2020transformers,choromanski2021rethinking,xiong2021nystromformer,lu2021soft,bolya2022hydra,cai2022efficientvit,zhen2022cosformer,liu2022ecoformer}.
Despite their promise in alleviating ViTs' complexity and thus inference runtime, both the local and linear attention compromise ViTs' performance due to the lack of capabilities to capture global or local context.
To marry the best of both worlds, we advocate training ViTs with both (1) efficient but less powerful linear attention, i.e., without the high-order residuals in angular kernel expansion, and (2) powerful yet costly softmax-based masked attention.
The latter helps approximate high-order residuals at the early training stage while being dropped during inference, based on an assumption that the remaining networks can gradually learn the
high-order components at the later training stage~\cite{xu2019frequency}.
This concept resembles the ``castling'' move in chess when two pieces are moved at once.
%
While it sounds promising, 
there are still two challenges to achieve this.
\textit{\uline{First}}, existing linear attention modules still underperform their vanilla softmax-based counterparts. Therefore, a better linear attention is crucial for the final performance. We find that angular kernels perform equally as softmax-based attentions in terms of similarity measurements. While they still suffer from a quadratic complexity, they can be divided into linear terms and high-order residuals. The challenge is \textit{how to construct ViTs with only the linear terms}.
%
\textit{\uline{Second}}, doing so would require that the trained ViTs merely rely on the linear terms towards the end of training, which would call for an approximation of the above high-order residuals.
The challenge is that \textit{how we can resort to costly but powerful modules to approximate high-order residuals during training but does not incur extra inference cost}.


In this work, we develop techniques to tackle those challenges, and make the following contributions:


\begin{itemize}
\itemsep -0.3\parsep

    
    
    
    \vspace{-0.3em}
    \item 
    We propose a framework called \textbf{Castling-ViT}, which trains ViTs using both linear-angular attention and masked softmax-based quadratic attention, but then switches to having only linear-angular attentions during ViT inference to save computational costs.
    

    \item We develop a new linear-angular attention leveraging angular kernels to close the accuracy gap between linear attention and softmax-based attention. 
    It expands angular kernels where linear terms are kept while complex high-order residuals are approximated.
    
    
    \item We use two parameterized modules to approximate the high-order residuals above: a depthwise convolution and an auxiliary masked softmax-based attention, where the latter's attention masks are regularized to gradually become zeros to avoid inference overhead.
    
    \item We conduct extensive experiments to validate the effectiveness of the proposed Castling-ViT. Results on classification, detection, and segmentation tasks consistently demonstrate its superior \hr{performance ($\uparrow$1.8\% top-1 accuracy or $\uparrow$1.2 mAP) or efficiency (40\% MACs savings)} over state-of-the-art (SOTA) CNNs and ViTs.
    
\end{itemize}

%% file: sections/2-Related_works.tex
  \vspace{-0.5em}
\section{Related Works}
\label{sec:related_works}
  \vspace{-0.3em}
\textbf{Vision Transformers (ViTs).}
ViT \cite{vit} beats CNNs with a simple encoder-only transformer architecture taking the splitted non-overlapped image patches as sequential inputs, but relies on costly pretraining on a huge JFT-300M dataset~\cite{chen2017revisiting}.
Later, DeiT~\cite{DeiT} and T2T-ViT \cite{yuan2021tokens} leverage an improved ViT training recipe or enhanced tokenization mechanism to achieve a comparable accuracy without the necessity of costly pretraining.
To further improve ViTs' achievable accuracy-efficiency tradeoffs,
CrossViT~\cite{chen2021crossvit}, PiT~\cite{heo2021rethinking}, PVT~\cite{wang2021pyramid}, MViT~\cite{fan2021multiscale} and Swin-Transformer~\cite{liu2021swin} propose a pyramid-like architecture, which is commonly used in CNNs~\cite{Dai2020FBNetV3JA,howard2019searching,xiong2020mobiledets}; DynamicViT~\cite{rao2021dynamicvit}, A-ViT~\cite{yin2022vit}, ToME~\cite{bolya2022token}, and MIA-Former \cite{yu2022mia} propose to adaptively identify and remove unnecessary input tokens for saving computational costs.
With the goal of deploying ViTs in resource-constrained devices, various efficient ViT architectures have been proposed
\cite{graham2021levit,li2022efficientformer,cai2022efficientvit,mehta2021mobilevit}. For example, LeViT~\cite{graham2021levit}, CvT\cite{wu2021cvt}, and MobileViT~\cite{mehta2021mobilevit} adopt more efficient self-attention implementation or incorporate convolutional feature extraction blocks into their early layers;
EfficientFormer \cite{li2022efficientformer} further enables pure ViTs to run as fast as MobileNets.
In contrast, our Castling-ViT explores whether ViTs can learn both global and local features while still being efficient at runtime. Also, we target
a generic linear-angular attention that can serve as a drop-in replacement for all kinds of ViT architectures and thus is orthogonal to new ViT architecture designs.

\textbf{Efficient ViT Variants.}
As commonly recognized, ViTs rely heavily on their self-attention module which is however costly due to its quadratic computational complexity with the total number of input tokens~\cite{zhu2021long,katharopoulos2020transformers}.
To make the self-attention module more efficient, a surge of linear attention works have been proposed and can be roughly categorized into two groups: local attention~\cite{liu2021swin,arar2022learned, wang2020linformer,tu2022maxvit} or kernel-based linear attention~\cite{katharopoulos2020transformers,choromanski2021rethinking,xiong2021nystromformer,lu2021soft,bolya2022hydra,cai2022efficientvit,zhen2022cosformer,liu2022ecoformer,ali2021xcit}.
For kernel-based linear attention, common designs approximate the softmax function~\cite{choromanski2021rethinking,katharopoulos2020transformers,bolya2022hydra} or the full self-attention matrix~\cite{xiong2021nystromformer,lu2021soft} with orthogonal features or kernel embeddings, then the computation order can be changed from $\mathbf{(QK)V}$ to $\mathbf{Q(KV)}$.
For example, \cite{katharopoulos2020transformers} and \cite{cai2022efficientvit} decompose the exponential terms in softmax-based attention into kernel functions and exchange the computation order.
Despite their decent performance, currently kernel-based linear attention in general underperform the softmax-based attention.
Recent works~\cite{chen2021scatterbrain,VITALITY} also unify low rank approximated and sparse attention (can also be dropped at inference) to improve ViTs' accuracy-efficiency tradeoffs.
Different from the above works, we explore from a new perspective by taking spectral angles into consideration when measuring the similarities among tokens, resulting in linear-angular attention that can achieve comparable or even better performance than softmax-based attention.
More efficient ViT variants are supplied to the Appendix.

ViTs have also been used as the backbones for downstream tasks, e.g., detection~\cite{zhu2020deformable,PVT_ICCV,li2022mvitv2} and segmentation~\cite{qin2022pyramid,cheng2022masked}. We supply more literature review to Appendix.

%% file: sections/3-Methods.tex
\vspace{-1em}
\section{The Proposed Methods}
\label{sec:methods}



\subsection{Preliminary of Self-Attention}
\label{sec:preliminary}
  \vspace{-0.3em}
\textbf{Self-Attention.}
Self-attention module is a core component of 
the Transformer
\cite{transformer,vit}, and usually consists of multiple heads.
Each head captures global-context information by measuring pairwise correlations among all $N$ tokens ($N$ denotes the total number of tokens) as defined below: 
\begin{equation} \label{equ:attn_op}
    \begin{split}
    \text{H}_{t}^{k} = 
    \sum_{i=1}^{N} \frac{{\tt exp}(\mathbf{Q}_t \cdot \mathbf{K}_i^T / \sqrt{d})}{\sum_{j=1}^{N} {\tt exp}(\mathbf{Q}_t \cdot \mathbf{K}_j^T / \sqrt{d})} \cdot \mathbf{V}_i,
    \end{split}
\end{equation}
where $t \in \{1, \cdots, N\}, k \in \{1, \cdots, M\}$, $\text{H}_t^k$ refers to the $t$-th row of the $m$-th head's attention matrix $\text{H}$. 
$\mathbf{Q}_t, \mathbf{K}_t, \mathbf{V}_t \in \mathbf{R}^d$,
are the query, key, and value vectors obtained by linearly projecting the input $\mathbf{X_t} \in \mathbb{R}^{D}$ with three learnable weight matrices $W^Q, W^K, W^V \in \mathbb{R}^{D \times d}$.
The attention head first computes the inner product between the query-key pairs, then scales the product results to stabilize the training and uses ${\tt Softmax}$ to normalize the resulting attention scores, and finally computes a weighted sum over all value vectors.
Outputs from all attention heads are then concatenated together before a final linear projection with learnable weights.
Note that ${\tt exp}(\cdot)$ denotes an exponential function.
Computing Eq. (\ref{equ:attn_op}) has a quadratic complexity of $\mathcal{O}(N^2)$.

\textbf{Kernel-based Linear Attention.} The core idea of linear attention \cite{katharopoulos2020transformers,wang2020linformer,shen2021efficient} is to decompose the similarity measurement function into separate kernel embeddings, i.e., $\text{Sim}(\mathbf{Q}, \mathbf{K}) \approx \phi(\mathbf{Q}) \phi(\mathbf{K})^T$, so that we can change the computation order to $\phi(\mathbf{Q}) (\phi(\mathbf{K})^T \mathbf{V})$ based on the associative property of matrix multiplication. In this way, the attention complexity is quadratic to the feature dimension $d$ instead of the token length $N$.
One straightforward implementation of linear attention is to use Gaussian RBF kernels to measure the similarity, which can serve as an unbiased estimation of $\text{exp}(\langle \cdot, \cdot \rangle)$ in Eq. (\ref{equ:attn_op}):
\begin{equation} \label{equ:rbf_kernel}
    \begin{split}
    \text{exp}(\frac{\langle \mathbf{x}, \mathbf{y} \rangle}{\sigma^2}) =
    \underbrace{
    \text{exp}(\frac{-\| \mathbf{x} - \mathbf{y} \|^2}{2\sigma^2})
    }_\textrm{\footnotesize Gaussian RBF Kernel}
    \text{exp}(\frac{\| \mathbf{x} \|^2 + \| \mathbf{y} \|^2}{2\sigma^2})
    \end{split}
\end{equation}
where $\langle \cdot , \cdot \rangle$ denotes the inner product operator. According to \cite{rahimi2007random}, we can induce a function $\phi(\cdot)$ to approximate the Gaussian RBF kernel, mapping the input space to the feature space. 
Assuming that both $\mathbf{Q}$ and $\mathbf{K}$ are normalized as unit row vectors along the feature dimension, then the attention formula can be approximated by:
\begin{equation} \label{equ:linear_attn}
    \begin{split}
    \text{H}_{t}^{k}
    & = \frac{\phi(\mathbf{Q}_t) \cdot \sum_{i=1}^{N} (\phi(\mathbf{K}_i)^T \cdot \mathbf{V}_i)}{\phi(\mathbf{Q}_t) \cdot \sum_{j=1}^{N} \phi(\mathbf{K}_j^T)}
    \end{split}
\end{equation}

While doing so can alleviate the attention complexity to become linear w.r.t. the token length $N$, it often causes a nontrivial accuracy drop as compared to the corresponding ViTs with vanilla softmax-based attention \cite{cai2022efficientvit,yuan2021tokens,choromanski2020rethinking}.

\begin{figure}[t]
    \centering
    \includegraphics[width=\linewidth]{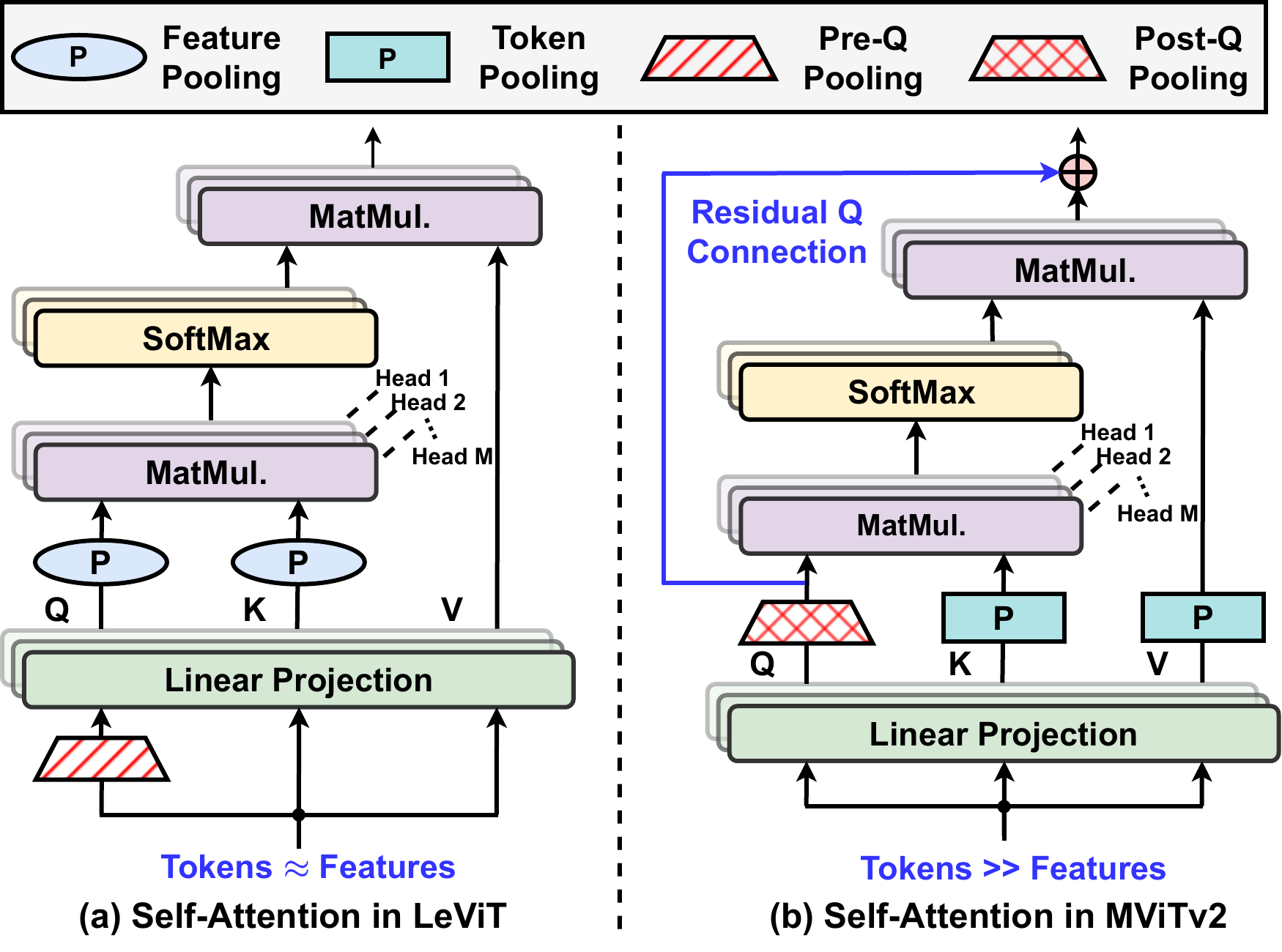}
    \vspace{-2em}
    \caption{Illustration of self-attention used in LeViT and MViTv2.}
    \label{fig:revisit}
    \vspace{-1.6em}
\end{figure}

\textbf{Revisit Attention Designs.} 
Recent ViTs have achieved a low complexity for classification tasks (e.g., LeViT~\cite{graham2021levit}) but still be costly for downstream tasks due to high input resolutions, i.e., the number of tokens (e.g., MViTv2~\cite{li2022mvitv2}). Therefore, one dilemma is the tradeoff between the model efficiency and generalizability.
As illustrated in Fig. \ref{fig:revisit}, LeViT shrinks the feature dimension to be more efficient on low-resolution tasks but cannot be well generalized to downstream tasks, \textcolor{black}{due to the resulting (1) feature bottlenecks, i.e., insufficient feature dimensions, and (2) fixed attention biases. On the other hand, MViTv2 performs token pooling for better fitting high-resolution tasks but is less effective for low-resolution due to the caused token bottlenecks, i.e., insufficient number of tokens.
Our ablation studies in Table \ref{tab:revisit_table} show that (1) for ImageNet classification, both token and feature pooling can lead to accuracy drops in ViTs; and (2) for downsampling layers, post-$\mathbf{Q}$ pooling (i.e., pooling after linear projection) together with residual connections in MViTv2 performs better than pre-$\mathbf{Q}$ pooling (i.e., pooling during linear projection) in LeViT.}
Therefore, we develop Castling-LeViTs on top of attention with merely post-$\mathbf{Q}$ pooling and residual connections.


\vspace{-0.1em}
\subsection{The Proposed Castling-ViT Framework}
\label{sec:castling-vit}
\vspace{-0.2em}

\textbf{Castling-ViT Overview.} Fig. \ref{fig:castling_vit} illustrates an overview of the proposed Castling-ViT, which 
makes linear attention more powerful than previous designs
while still being efficient during inference. In particular, we propose (1) a novel kernel-based linear-angular attention from the spectral angle perspective to close the accuracy gap between linear attention and softmax-based attention; and (2) a training augmentation method that leverages softmax-based attention as an auxiliary branch to assist the linear-angular attention only during ViT training. Note that a mask is applied to the auxiliary branch to manifest linear-angular attention and drop the auxiliary branch. 

\input{tables/revisit}

\vspace{-0.7em}
\subsubsection{Linear-Angular Attention}
\label{sec:linear_angular}
\vspace{-0.2em}

\textbf{Angular Kernel.} 
In addition to the previously adopted polynomial, exponential, or RBF kernel that focuses on spatial similarity measurements \cite{katharopoulos2020transformers,liu2022ecoformer}, we propose to consider measuring spectral similarity via angular kernel as an alternative to existing softmax-based attentions, leading to similar or better performance since it additionally takes into consideration the nature of spectral characteristics, e.g., the spectral angle as a distance measurement function \cite{keshava2004distance,honeine2010angular}.
Such a spectral angle between two vectors is defined as:
\begin{equation} \label{equ:angular}
    \begin{split}
    \theta(\mathbf{x}_i, \mathbf{x}_j) = \text{arccos}\left( \frac{\langle \mathbf{x}_i, \mathbf{x}_j \rangle}{\|\mathbf{x}_i\| \cdot \|\mathbf{x}_j\|} \right),
    \end{split}
\end{equation}
where $\|\cdot\|$ is the Euclidean distance and $\langle \cdot, \cdot \rangle$ is the inner product. The output range of $\theta$ is $[0, \pi]$.
Such an angle can be used as a distance. In our design, we define the angular kernel as a similarity measurement function between the queries $\mathbf{Q}$ and keys $\mathbf{K}$ as:
\begin{equation} \label{equ:angular_kernel}
    \begin{split}
    \textrm{Sim}(\mathbf{Q}_i, \mathbf{K}_j) = 1 - \frac{1}{\pi} \cdot \theta(\mathbf{Q}_i, \mathbf{K}_j),
    \end{split}
\end{equation}
and the output range is thus $[0, 1]$. With more aligned $\mathbf{Q}_i$ and $\mathbf{K}_j$, $\theta$ is closer to 0 and thus the similarity is closer to 1; In contrast, if $\mathbf{Q}_i$ and $\mathbf{K}_j$ have opposite features, $\theta$ is closer to $\pi$ and thus the similarity is closer to 0.

\begin{figure}[t]
    \centering
    \includegraphics[width=0.8\linewidth]{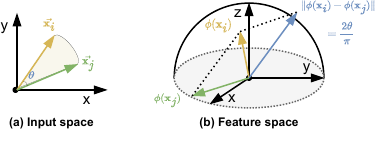}
    \vspace{-2em}
\caption{(a) A simple 2D input space and (b) the corresponding 3D feature space after applying the angular kernel to the 2D input data. Note that real input data/features are of higher dimensions.}
    \label{fig:angular_vis}
    \vspace{-1.5em}
\end{figure}

\textbf{Properties of Angular Kernel and Its Feature Space.}
One property of our angular kernel is that 
replacing the similarity measurement in self-attention with such a kernel provides an efficient way to implicitly map the input data to a high (even infinite after expansion) dimensional feature space~\cite{honeine2010angular}, where the distances/angles are calculated based on a rich feature structure.
Let $\phi(\cdot)$ denotes the implicit map induced by this kernel, the norm of mapped input data is:
\begin{equation} \label{equ:norm}
    \begin{split}
    \|\phi(\mathbf{x}_i)\|^2 = \textrm{Sim}(\mathbf{x}_i, \mathbf{x}_i) = 1 - \frac{1}{\pi} \cdot 0 = 1,
    \end{split}
\end{equation}
which means that all data in the input space are mapped onto the sphere of radius 1 in the feature space. Also, the distance between two input features is given by:
\begin{equation} \label{equ:distance}
    \begin{split}
    \|\phi(\mathbf{x}_i) & - \phi(\mathbf{x}_j)\|^2 = 1 + 1 - 2 \cdot \phi(\mathbf{x}_i)\phi(\mathbf{x}_j)^T \\
    & = 2 \cdot (1 - \textrm{Sim}(\mathbf{x}_i, \mathbf{x}_j)) = \frac{2}{\pi} \cdot \theta(\mathbf{x}_i, \mathbf{x}_j).
    \end{split}
\end{equation}
That is, the square (Euclid) distance and the spectral angle is positively correlated and the distance range is $[0, 2]$.
As shown in Fig. \ref{fig:angular_vis}, the angles in the input space is correlated to the feature distance after applying our angular kernel.

\textbf{Expansion of Angular Kernel.}
A natural following question is how to incorporate the above angular kernel for designing linear attention given its quadratic complexity w.r.t. the input token length. Recalling from trigonometric identities and the expansion of the \textrm{arccos} function into an infinite series, we reformulate the similarity function as:
\begin{equation} \label{equ:angular_kernel_expansion}
    \begin{split}
    \textrm{Sim}(&\mathbf{Q}_i, \mathbf{K}_j) =
    1 - \frac{1}{\pi} \cdot \left( \frac{\pi}{2} - \textrm{arcsin} \left( \mathbf{Q}_i \cdot \mathbf{K}_j^T \right)  \right) \\
    %
    %
    & = 
    \underbrace{
    \frac{1}{2} + \frac{1}{\pi} \cdot \left( \mathbf{Q}_i \cdot \mathbf{K}_j^T \right)
    }_\textrm{\footnotesize Linear-angular terms} \\
    & + 
    \underbrace{
    \frac{1}{\pi} \cdot \sum_{k=1}^{\infty} \frac{(2k)!}{2^{2k} (k!)^2 (2k+1)} \left( \mathbf{Q}_i \cdot \mathbf{K}_j^T \right)^{2k+1}
    }_\textrm{\footnotesize High-order residuals when $k \geq 1$ (non-linear)},
    \end{split}
\end{equation}
where $\mathbf{Q}_i \cdot \mathbf{K}_j^T$ denotes the normalized linear kernel function $\langle \mathbf{Q}_i, \mathbf{K}_j \rangle / \|\mathbf{Q}_i\| \cdot \|\mathbf{K}_j\|$, which is equivalent to the inner product if $\mathbf{Q}_i$ and $\mathbf{K}_j$ are unit vectors.
We see that the first linear-angular terms can be directly used as the similarity measurement in linear attention, while the remaining higher-order terms of an infinite series introduce a much higher complexity. As such, we propose to adopt a relaxation to approximate it.

\begin{figure}[t]
    \centering
    \includegraphics[width=\linewidth]{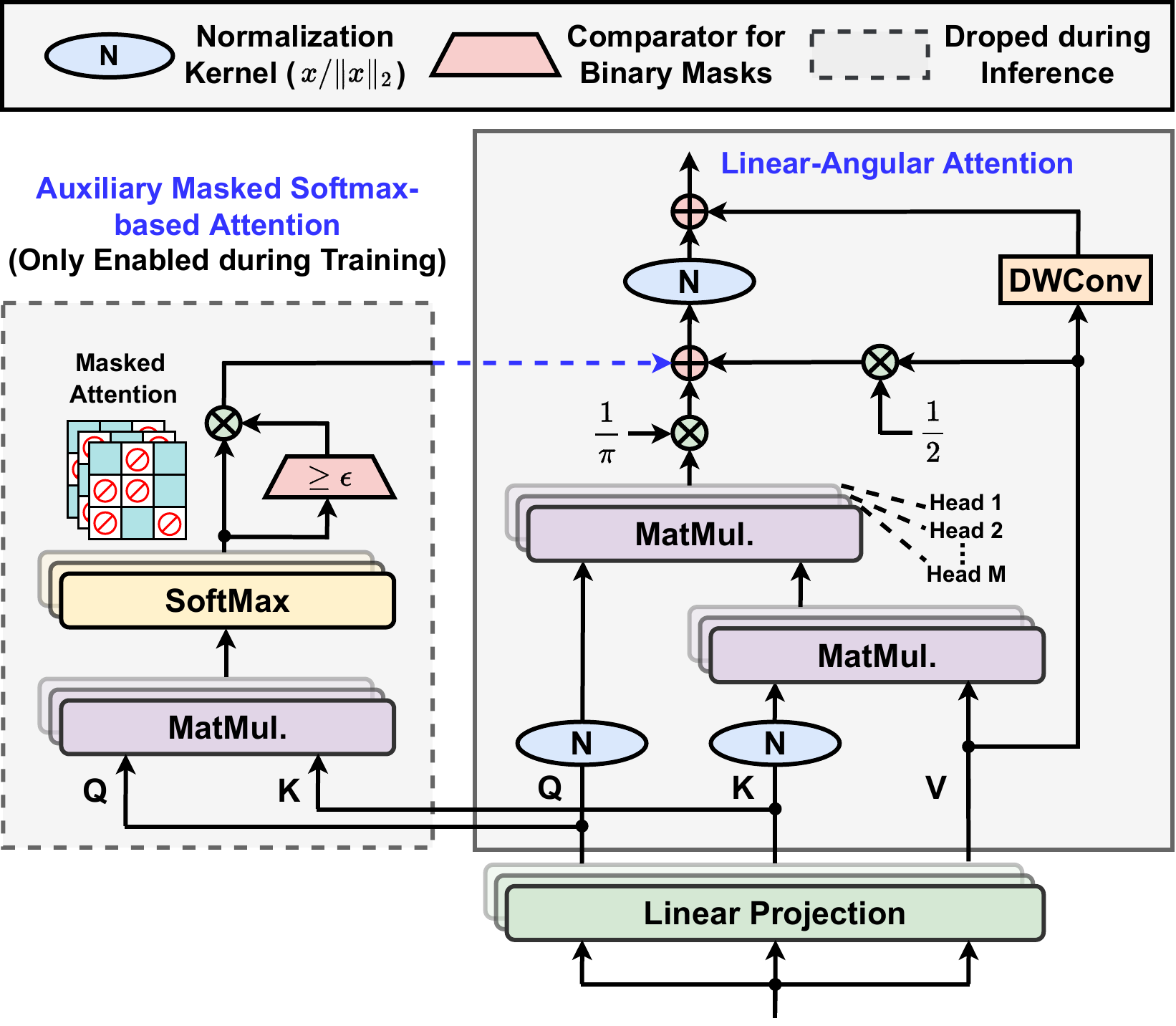}
    \vspace{-2em}
    \caption{The visualization of our Castling-ViT with both linear-angular attention and auxiliary softmax-based self-attention.}
    \label{fig:castling_vit}
    \vspace{-1em}
\end{figure}

\textbf{Linear-Angular Attention.} 
To construction our linear-angular attention,
we leverage parametrized DNN modules to approximate the expectation of the high-order residuals, i.e., $\mathbf{M}_{ij} = \mathbb{E}_{\mathbf{Q}_i \sim p_{{}_{Q_i}}, \mathbf{K}_j \sim p_{{}_{K_j}}} \left(\sum_{k=1}^\infty \alpha_k (\mathbf{Q}_i \cdot \mathbf{K}_j^T)^{2k+1} \right)$, where $\alpha_k$ is the coefficient for the $k$-th order term in Eq. (\ref{equ:angular_kernel_expansion}).
The key question is \textit{how to design such DNN modules}.
Kim et al. \cite{kim2021rethinking} conducted an analysis on average attention weights of ViT models, which show a strong inductive bias to attend to neighboring tokens. To capture this, we introduce a learnable depthwise convolution (DWConv) module which is applied on all the value tokens, so $\mathbf{M}_{ij}$ by-design attends to nearby tokens, as illustrated in Fig. \ref{fig:castling_vit}.
Second, DWConv is limited by its receptive field~\cite{moody1988learning} while the averaged attention map also has a small number of off-diagonal scores to connect to nonadjacent tokens~\cite{kim2021rethinking}. To capture that, we adopt a sparse softmax-based attention as described in Section \ref{sec:castling}. In practice, we found that such a sparse softmax attention tends to converge to all zeros after adding a simple threshold regularization. We provide more explanation and visualization in Sec. \ref{sec:castling} and Sec. \ref{sec:discussion}.


In this way, our attention module can be formulated as:
\begin{equation} \label{equ:linear_angular_attention}
    \begin{split}
    \textrm{H} & = \textrm{Sim}(\mathbf{Q}, \mathbf{K}) \cdot \mathbf{V} 
    \approx \frac{1}{2} \cdot \mathbf{V} + \frac{1}{\pi} \cdot \mathbf{Q} \cdot (\mathbf{K}^T \cdot \mathbf{V})  \\ &  + \mathbf{M}_{DW} \cdot \mathbf{V} + \mathbf{M}_{Sparse Attn},
    \end{split}
\end{equation}
where $1/\pi \cdot Q \cdot (K^T \cdot V)$ is the linear term with $\mathcal{O}(N)$ complexity, $\mathbf{M}_{DW} \cdot V$ is the matrix form of DWConv, and  $\mathbf{M}_{Sparse Attn}$ is the \hry{normalized} sparse softmax attention. The overall complexity to compute 
Eq. (\ref{equ:linear_angular_attention}) 
is linear to the input token length, where the MACs of DWConv is also negligible (e.g., $<$ 1\% of the total MACs).
Also, normalizations ($\nicefrac{\mathbf{x}}{\|\mathbf{x}\|_2}$) are inserted to the $\mathbf{Q}/\mathbf{K}$ \hry{and sparse attention} branches to help the similarity measurement following \cite{bolya2022hydra}.

\vspace{-1em}
\subsubsection{Switch Towards Linear-Angular Attention}
\label{sec:castling}
\vspace{-0.3em}

Recall that we add a sparse softmax-based attention as an auxiliary branch in the linear-angular attention to help approximate the high-order residuals.
Such a costly attention can be potentially dropped without hurting the inference accuracy, drawing inspiration from ~\cite{xu2019frequency} that the remaining network can gradually learn the high-order/frequency components at the later training stage.
Next, we explain how it is constructed and dropped.

\textbf{Sparse Training Augmentation.} 
As illustrated in Fig. \ref{fig:castling_vit}, we adopt a masked softmax-based attention as an auxiliary branch to augment ViT training. 
In particular, we first use a comparator with a predefined threshold $\epsilon$ to generate a binary mask, where attentions greater than $\epsilon$ are set to 1 and 0 otherwise.
These masks are then applied to the attention maps to generate masked attention maps that will be summed up with our linear-angular attention together as the final attention. The resulting sparse attention is given by:
\vspace{-0.2em}
\begin{equation} \label{equ:sparse}
    \begin{split}
    \mathbf{M}_{Sparse Attn}(\mathbf{Q, K}) = \tt{Mask}_{\epsilon} ({\tt Softmax}(\mathbf{Q}\cdot \mathbf{K}^T)).
    \end{split}
\end{equation}
\vspace{-0.2em}
%
where $\tt{Mask}_{\epsilon}(x) = x \text{ if } x > \epsilon \text{ else } 0$, acting as an element-wise threshold function. As such, the sparse attention captures the higher attention scores (i.e., strong local features), and can potentially complement our linear-angular attention by supplementing the missing higher-order terms.
Such an assumption aligns well with recent findings that (1) low-rank and sparse approximations complement each other \cite{chen2021scatterbrain,VITALITY,kim2021rethinking}; and (2) linear attention lacks local feature extraction capabilities over its softmax-based counterpart \cite{cai2022efficientvit}. 


\textbf{Castling During ViT Inference.}
As we are targeting efficient ViT inference, it is desired to reduce or completely remove the costly softmax-based attentions while only keeping our linear-angular attention at runtime, i.e., performing castling. In our experiments, we found that under the sparsity regularization above, the softmax attention naturally converges to all zeros as the training progresses. We consider both fixed and dynamic schedules for $\epsilon$ and find that our Castling-ViT is not sensitive to neither the threshold value nor the threshold schedule for a given task. Given a fixed mask threshold (e.g., $\epsilon = 0.02$ in image classification experiments), the masks become all zeros at latter training stages and thus the auxiliary branch can be removed without hurting the model accuracy.
We supply the visualization of the mask-evolving trajectory and our conjecture for understanding such a phenomenon in Sec. \ref{sec:discussion}.

%% file: tables/revisit.tex
\begin{table}[!t]
    \centering
    \caption{Analyze the attention design on ImageNet classification.}
    \vspace{-0.8em}
    \resizebox{\linewidth}{!}{
      \begin{tabular}{cc|cc|c|c|c}
      \hline
      \multicolumn{4}{c|}{\textbf{Pooling}} & \multirow{2}[1]{*}{\textbf{Residual} $\mathbf{Q}$} & \multirow{2}[1]{*}{\textbf{\hry{MACs} (M)}} & \multirow{2}[0]{*}{\tabincell{c}{\textbf{Top-1}\\\textbf{Accuracy (\%)}}} \\
    \cline{1-4}  \textbf{Token} & \textbf{Feat.} & \textbf{Pre-$\mathbf{Q}$} & \textbf{Post-$\mathbf{Q}$} &   &   &  \\
      \hline
      \hline
      \Checkmark &   & \Checkmark &   &   & 685 & 78.00  \\
        & \Checkmark & \Checkmark &   &   & 661 & 78.19 \\
        &   & \Checkmark &   &   & 736 & 79.11  \\
      \hline
      \Checkmark &   &   & \Checkmark & \Checkmark & 771 & 78.73 \\
        & \Checkmark &   & \Checkmark & \Checkmark & 763 & 77.23 \\
        &   &   & \Checkmark & \Checkmark & 838 & 80.05 \\
      \hline
      \end{tabular}%
    }
    \label{tab:revisit_table}
    \vspace{-1.em}
\end{table}

%% file: sections/4-Experiments.tex
\vspace{-0.3em}
\section{Experiments}
\label{sec:exps}


\vspace{-0.3em}
\subsection{Experiment Settings}
\vspace{-0.3em}

\textbf{Tasks, Datasets, and Models.}
\underline{\textit{Tasks and Datasets.}} We consider three benchmark datasets and three representative vision tasks to demonstrate the superiority of the proposed Castling-ViT, including image classification on ImageNet dataset \cite{deng2009imagenet} with 1.2 million training and 50K validation images; 
Object detection on COCO dataset \cite{lin2014microsoft} with 118K training and 5K validation images;
Semantic segmentation on ADE20K dataset \cite{zhou2017scene} with 20K/2K/3K images for training, validation, and testing, respectively.
\underline{\textit{Models.}} We apply our proposed Castling-ViT idea on top of various models. For the classification task, we consider LeViT \cite{graham2021levit}, MViTv2 \cite{li2022mvitv2}, and DeiT \cite{DeiT}; 
For the detection task, we consider models with efficient ViT backbones (e.g., PicoDet \cite{yu2021pp} with modified ESNet and LCNet backbones \cite{cui2021pp} with transformer blocks);
For the segmentation task, we consider Mask2former \cite{cheng2022masked} with ViT-Base backbone.

\input{tables/imagnet}

\textbf{Training Settings.}
\underline{\textit{For the classification task,}} we use a SGD optimizer with 0.9 momentum and $2\times 10^{-5}$ weight decay to train ViTs for 1000 epochs using 64 V100 GPUs, with each card having 64 (LeViT) or 32 (MViTv2/DeiT) batch sizes.
The learning rate is 2.0 with first 11 epochs warm-up starting from 0.01 and decays by a factor of 0.9875 per epoch \cite{wu2021fbnetv5}. Also, we use the distillation \cite{DeiT} based on a teacher model with a 85.5\% accuracy.
\underline{\textit{For the detection task,}} we adopt SGD optimizer with momentum 0.9 and weight decay 4e-5 to train models on COCO.
All models are trained on 8 V100 GPUs with each card having 80 batch sizes following PicoDet's training recipe~\cite{yu2021pp}. Also, we follow LeViT's training recipe \cite{graham2021levit} to pretrain backbones on ImageNet;
\underline{\textit{For the segmentation task,}} we follow Mask2former's training recipe \cite{cheng2022masked} to train models on ADE20K, where ViT backbones are pretrained following MAE \cite{he2022masked} if specified.

\textbf{Baselines and Evaluation Metrics.}
\underline{\textit{Baselines.}} For the classification task, we compare the proposed Castling-ViT with LeViT~\cite{graham2021levit}, MviTv2~\cite{li2022mvitv2}, DeiT~\cite{DeiT}, Swin~\cite{liu2021swin}, CSWin~\cite{dong2022cswin}, PVT~\cite{wang2021pyramid}, etc.
For the detection task, we compare with FBNetV5~\cite{wu2021fbnetv5}, YOLOX~\cite{ge2021yolox}, YOLOv5, MobileDet~\cite{xiong2020mobiledets}, and EfficientDet~\cite{tan2020efficientdet}.
For the segmentation task, we compare with Mask2former~\cite{cheng2022masked} with ViT backbones.
\underline{\textit{Evaluation Metrics.}} We evaluate the Castling-ViT and all baselines in terms of accuracy-efficiency tradeoffs. Specifically, the accuracy metrics refer to top-1/5 accuracy for the classification task; AP, AP$^{50}$, AP$^{75}$ for the detection task (AP: average precision); mIoU, mAcc, and pAcc for the segmentation task (mIoU: mean intersection over union). For efficiency metrics, we compare the number of model parameters or inference FLOPs (or MACs).

\vspace{-0.2em}
\subsection{Castling-ViT over SOTA Baslines}
\vspace{-0.3em}

\begin{figure}[t]
    \centering
    \includegraphics[width=\linewidth]{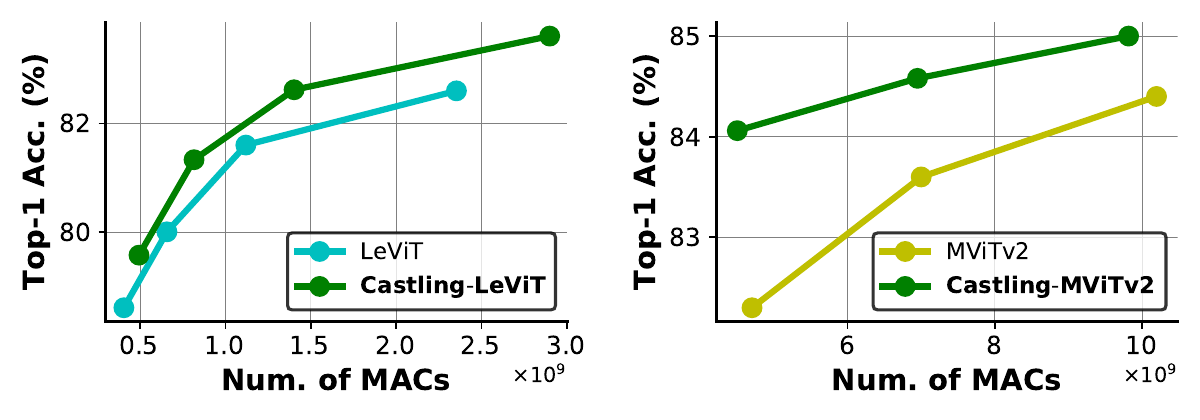}
    \vspace{-2em}
    \caption{Castling vs. baseline ViT on ImageNet.}
    \label{fig:imagenet}
    \vspace{-1.5em}
\end{figure}

\textbf{Image Classification.} To evaluate the effectiveness of our proposed techniques on the image classification task, we apply the proposed Castling-ViT idea to three typical or SOTA ViT architectures: DeiT~\cite{DeiT} as typical ViTs, LeViT~\cite{graham2021levit} as representative efficient ViTs, and MViTv2~\cite{li2022mvitv2} as representative hierarchical ViTs for downstream applications of high input resolutions, and compare their performance over baselines on ImageNet.
As shown in Tab. \ref{tab:imagenet}, the comparison across a large MACs (or FLOPs) range starting from 0.4G to 17G. We categorize ViT models into four regimes: $<$1G, 1$\sim$3G, 3$\sim$10G, and $>$10G, and select baselines in each regime to benchmark separately for the clarity purpose.
Castling-ViTs consistently perform better than all baselines across various MACs ranges in terms of the accuracy-efficiency tradeoff. For example, Castling-LeViT achieves 82.6\% top-1 accuracy with only 1.40G MACs while LeViT requires 2.35G FLOPs instead, i.e., \textbf{$\downarrow$40\% MACs};
On the other hand, under comparable MACs, Castling-MViTv2 achieves \hr{84.1\%} accuracy vs. \hr{MViTv2} with \hr{82.3\%} accuracy instead, i.e., \textbf{$\uparrow$\hr{1.8}\% top-1 accuracy}.
Overall, Castling-ViT achieves an improved accuracy of \textbf{\hr{$\uparrow$0.5\% $\sim$ $\uparrow$6.6\%}}, \textbf{\hr{$\uparrow$1.0\% $\sim$ $\uparrow$8.1\%}}, \textbf{\hr{$\uparrow$1.0\% $\sim$ $\uparrow$4.1\%}}, and \textbf{\hr{$\uparrow$0.6\% $\sim$ $\uparrow$2.6\%}} over baselines under$<$1G MACs, 1$\sim$3G MACs, 3$\sim$10G MACs, and $>$10G MACs, respectively. Note that we calculate improvements under comparable MACs.
In addition to the overall comparison, we also visualize the apple-to-apple benchmark, e.g., Castling-LeViT vs. LeViT, to validate the effectiveness of proposed techniques.
As shown in Fig. \ref{fig:imagenet}, Castling-LeViT/MViTv2 achieves \textbf{\hr{$\downarrow$25.7\% $\sim$ $\downarrow$55.3\%}} MACs reductions under comparable accuracies or offers a comparable or better accuracy (\textbf{\hr{$\uparrow$0.6\% $\sim$ $\uparrow$1.8\%}}) under comparable MACs over corresponding LeViT/MViTv2 baselines.

\input{tables/coco_small}

\textbf{Object Detection.}
We also extend the Castling-ViT to the downstream object detection task and compare it with previous efficient detectors on COCO dataset to evaluate its efficacy.
Specifically, we construct the detector with modified ESNet~\cite{yu2021pp} or LCNet~\cite{cui2021pp} (replace the last one or two stages with transformer blocks) as backbones and follow PicoDet~\cite{yu2021pp}'s detection head design as well as their training recipe.
Fig. \ref{fig:overall_comp} and Tab. \ref{tab:coco_small} show the overall comparison between the proposed Castling-ViT and other baselines. We can see that our Castling-ViT consistently achieves better accuracy-efficiency tradeoffs, leading to \textbf{\hr{$\uparrow$6.0, $\uparrow$2.2 $\sim$ $\uparrow$2.3, $\uparrow$4.0 $\sim$ $\uparrow$5.9, $\uparrow$3.1 $\sim$ $\uparrow$4.0 mAP}} improvements as compared to YOLOv5, YOLOX~\cite{ge2021yolox}, MobileDet~\cite{xiong2020mobiledets}, and FBNetv5~\cite{wu2021fbnetv5}, respectively, under comparable or even less MACs.
As for EfficientDet~\cite{tan2020efficientdet}, our method achieves comparable accuracy-efficiency trade-offs.
Apart from the overall comparison with baselines, we also provide the apple-to-apple comparison with detectors with softmax-based attention, we supply the comparison results and Castling-ViT's breakdown analysis to Sec. \ref{sec:ablation_castling_vit}.
This set of experiments validate the effectiveness of the proposed Castling-ViT for servering as efficient detector backbones (e.g., $\leq$3G MACs) in the object detection task.

\textbf{Semantic Segmentation.}
We further extend Castling-ViT to the semantic segmentation task to evaluate its effectiveness.
Specifically, we use ViT-Base as the backbone in Mask2former~\cite{cheng2022masked} framework to serve as our baseline and testbed. Then we build the Castling-ViT-Base as the backbone to benchmark on the ADE20K dataset.
As shown in Tab. \ref{tab:ade20k}, Mask2former with Castling-ViT backbone achieves \textbf{15\%} total MACs reductions and \textbf{19\%} backbone MACs reduction under comparable or slightly better mIoU, i.e., \textbf{$\uparrow$0.13\%} and \textbf{$\uparrow$0.52\%} without or with MAE pretraining on ImageNet~\cite{he2022masked}.
This set of experiments validate that our proposed techniques could be well generalized to various downstream tasks that require large input resolutions.

\input{tables/ade20k}


\vspace{-0.2em}
\subsection{Linear-Angular Attention over SOTA Baselines}
\vspace{-0.2em}

\input{tables/ablation_linear_attn}

We also conduct ablation studies among various kinds of kernels used in linear attention to evaluate the superiority of our proposed linear-angular kernel.
Also shown in Tab. \ref{tab:ablation_linear_attn}, we compare the angular kernel with five other commonly adopted kernels, results on three model and resolution settings consistently demonstrate that our proposed angular kernel helps achieve better mAP, e.g., $\uparrow$4.0\% $\sim$ $\uparrow$4.2\% over MC~\cite{VITALITY}, $\uparrow$1.0\% $\sim$ $\uparrow$1.2\% over Softmax~\cite{shen2021efficient}, $\uparrow$2.6\% $\sim$ $\uparrow$3.2\% over Cosine~\cite{bolya2022hydra}, $\uparrow$3.1\% $\sim$ $\uparrow$4.6\% and $\uparrow$0.2\% $\sim$ $\uparrow$1.2\% over ReLU-S and ReLU-E~\cite{cai2022efficientvit}, respectively.
This set of experiments validate the superiority of the proposed linear-angular kernels.

\vspace{-0.2em}
\subsection{Ablation Studies of Castling-ViT}
\label{sec:ablation_castling_vit}
\vspace{-0.2em}

We conduct ablation studies on Castling-ViT's linear-angular attention, added DWConv, as well as auxiliary masked softmax-based attention, as shown in Tab. \ref{tab:ablation_s_320} and Tab. \ref{tab:ablation_m_416}, where the experiments are performed on Castling-ViT-S-320$^{\star}$/M-416$^{\star}$, respectively, without pretraining on ImageNet. Note that here $^{\star}$ means that we use LCNet-ViT, i.e., replacing all layers in last two stages with transformer blocks, as backbones.
Results in three tables show detailed performance breakdown and consistently demonstrate that all components in our proposed Castling-ViT contribute to the final performance.
Specifically, linear-angular attention itself already achieves $\downarrow$0.1\% $\sim$ $\uparrow$0.5\% mAP improvements while leading to 10.1\% $\sim$ 15.3\% MACs reductions simultaneously.
Adding DWConv in the middle of MLP layers further leads to $\downarrow$0.1\% $\sim$ $\uparrow$0.6\% mAP improvements while only incur negligible MACs overhead.
Adding auxiliary masked softmax-based attention further increases the mAP by 0.2\% without incurring any overhead since it only assists the training while being removed during inference.
%
Note that although LCNet-ViT backbone leads to better mAP, it also introduces more MACs 
as compared to efficient convolutions.
As such, LCNet-ViT still slightly underperforms LCNet backbones in terms of accuracy-efficiency trade-offs, 
since we directly replace the CNN backbones with transformer blocks, whose architecture could not be optimal for ViTs. It remains to an open problem to build ViT based models that achieves higher accuracy-efficiency trade-off than pure ConvNet models.

\input{tables/ablation_s_320}

\input{tables/ablation_m_416}

\vspace{-0.3em}
\subsection{Discussion on the Auxiliary Branch}
\label{sec:discussion}
\vspace{-0.2em}

\begin{wrapfigure}{r}{0.22\textwidth}
    \vspace{-0.2in}
    \centering
    \includegraphics[width=0.22\textwidth]{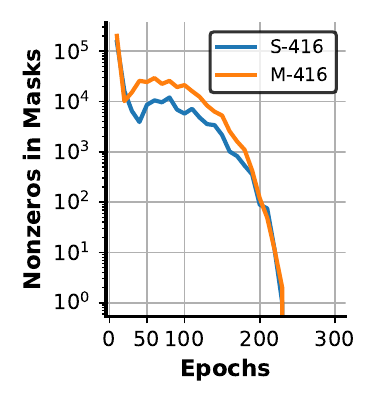}
    \vspace{-2.5em}
    \caption{Visualizing the trajectories of nonzeros in auxiliary masks during training.}
    \label{fig:nonzeros}
    \vspace{-1em}
\end{wrapfigure}
\textbf{Trajectory of Nonzeros in Masks.}
To further understand the effect of the auxiliary branch, we visualize the trajectories of nonzeros in masks of auxiliary softmax-based attention.
As shown in Fig. \ref{fig:nonzeros}, we count the nonzeros in the first attention layer's masks throughout the training of both Castling-ViT-S-416 and Castling-ViT-M-416 with the LCNet-ViT backbone on COCO.
we observe that the introduced auxiliary attention will only assist the training in the early or middle training stages and will gradually vanish towards all zeros in the later training stage, which is well aligned with the assumption in our Castling-ViT. 
This set of experiments validate the idea of performing ``castling'' in ViTs, i.e., drop the auxiliary branch without sacrificing performance.

\begin{figure}[t]
    \centering
    \includegraphics[width=\linewidth]{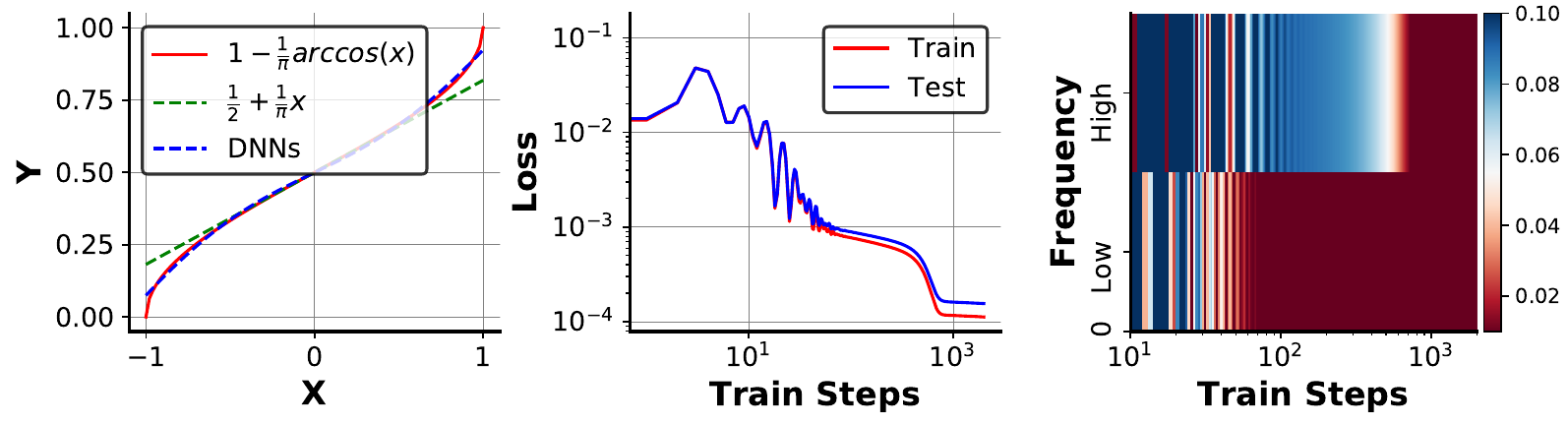}
    \vspace{-2em}
    \caption{\textbf{Left:} Visualizing the taget similarity function, linear-angular terms, and DNNs' fitting curve; \textbf{Middle:} Visualizing the loss trajectory during DNN training; \textbf{Right:} Visualizing the learning trajectory of both low and high frequency components.}
    \vspace{-1.6em}
    \label{fig:fitting}
\end{figure}

\textbf{Conjecture of Castling.}
We conduct a synthetic experiment to give an analogy for explaining the castling phenomenon.
In Fig. \ref{fig:fitting} (Left), we visualize the curve of (1) angular similarity function (denoted in \textcolor[RGB]{227,23,13}{Red}), (2) linear-angular terms (denoted in \textcolor[RGB]{34,139,34}{Green}), and (3) two-layer DNN's approximation.
We see that only keeping linear-angular terms leads to distortion,
while DNN is capable of learning the missing high-frequency parts,
whose loss trajectory is shown in Fig. \ref{fig:fitting} (Middle).
Also, the learning trajectory of frequency components is visualized in Fig. \ref{fig:fitting} (Right), following F-Principle~\cite{xu2019frequency}, where x-axis is training steps, y-axis categorizes both low and high frequency components, blue/red colors refer to large/small difference between the learned frequency components and target frequency components, respectively.
We observe that 
DNNs fit target functions from low to high frequencies during training.
%
It indicates that DWConv itself is not sufficient for approximating the high-order residual at early training stages.
Costly attention that contains high-order residuals is then desired to help training while being dropped at inference since the remaining networks can gradually learn the high-frequency components at later training stages~\cite{xu2019frequency}.

%% file: tables/imagnet.tex
\begin{table}[t]
  \centering
  \caption{Castling-ViT over SOTA baselines on ImageNet.}
  \vspace{-0.8em}
  \resizebox{\linewidth}{!}{
  \begin{tabular}{c|l|cc|cc}
  \hline
  \textbf{\tabincell{c}{MACs\\Ranges}} & \textbf{Models} & \textbf{\tabincell{c}{Params\\(M)}} & \textbf{\tabincell{c}{MACs\\(G)}} & \textbf{\tabincell{c}{Top-1\\Acc.(\%)}} & \textbf{\tabincell{c}{Top-5\\Acc.(\%)}} \\
  \hline
  \hline
  \multirow{8}[2]{*}{$<$1G} &        MobileNetV2-1.4~\cite{sandler2018mobilenetv2} & 6.9 & 0.58 & 74.7 & - \\
    & EfficientNet-B0~\cite{tan2019efficientnet} & 5.3 & 0.39 & 77.1 & 93.3 \\
    & EfficientNet-B1~\cite{tan2019efficientnet} & 7.8 & 0.70 & 79.1 & 94.4 \\
    & MobileViT-XS~\cite{mehta2021mobilevit} & 2.3 & 0.70 & 74.8 & 92.3 \\
    & LeViT-128~\cite{graham2021levit} & 9.2 & 0.41 & 78.6 & - \\
    & LeViT-192~\cite{graham2021levit} & 10.9 & 0.66 & 80.0 & - \\
    \rowcolor{gray}
    \cellcolor{white} & \textbf{Castling-LeViT-128} & 10.5 & 0.49 & \textbf{79.6} & \textbf{94.6} \\
    \rowcolor{gray}
    \cellcolor{white} & \textbf{Castling-LeViT-192} & 12.7 & 0.82 & \textbf{81.3} & \textbf{95.5} \\
  \hline
  \multirow{8}[2]{*}{1$\sim$3G} & EfficientNet-B3~\cite{tan2019efficientnet} & 12.0 & 1.80 & 81.6 & 95.7 \\
    & HRFormer-T~\cite{yuan2021hrformer} & 8.0 & 1.80 & 78.5 & - \\
    & DeiT-T~\cite{DeiT} & 5.6 & 1.25 & 74.5 & 91.9 \\
    & LeViT-256~\cite{graham2021levit} & 18.9 & 1.12 & 81.6 & - \\
    & LeViT-384~\cite{graham2021levit} & 39.1 & 2.35 & 82.6 & - \\
    & Caslting-DeiT-T & 5.64 & 1.18 & 76.0 & 92.5 \\
    \rowcolor{gray} 
    \cellcolor{white} & \textbf{Castling-LeViT-256} & 22.0 & 1.40 & \textbf{82.6} & \textbf{96.1} \\
    \rowcolor{gray}
    \cellcolor{white} & \textbf{Castling-LeViT-384} & 45.8 & 2.90 & \textbf{83.7} & \textbf{96.7} \\
  \hline
  \multirow{16}[2]{*}{3$\sim$10G} & RegNetY-4G~\cite{radosavovic2020designing} & 21.0 & 4.00 & 80.0 & - \\
    & RegNetY-8G~\cite{radosavovic2020designing} & 39.0 & 8.00 & 81.7 & - \\
    & EfficientNet-B4~\cite{tan2019efficientnet} & 19.0 & 4.20 & 82.9 & 96.4 \\
    & EfficientNet-B5~\cite{tan2019efficientnet} & 30.0 & 9.90 & 83.6 & 96.7 \\
    & Swin-T~\cite{liu2021swin} & 29.0 & 4.50 & 81.3 & 95.5 \\
    & Swin-S~\cite{liu2021swin} & 50.0 & 8.70 & 83.2 & 96.2 \\
    & CSWin-T~\cite{dong2022cswin} & 23.0 & 4.30 & 82.7 & - \\
    & CSWin-S~\cite{dong2022cswin} & 35.0 & 6.90 & 83.6 & - \\
    & DeiT-S~\cite{DeiT} & 21.9 & 4.60 & 81.2 & 95.4 \\
    & AutoFormer-S~\cite{chen2021autoformer} & 22.9 & 5.10 & 81.7 & 95.7 \\
    & PVTv2-V2~\cite{wang2022pvt} & 25.0 & 4.00 & 82.0 & - \\
    & MViTv2-T~\cite{li2022mvitv2} & 24.0 & 4.68 & 82.3 & - \\
    & MViTv2-S~\cite{li2022mvitv2} & 34.7 & 6.95 & 83.6 & - \\
    \rowcolor{gray}
    \cellcolor{white} & \textbf{Castling-MViTv2-T} & 24.1 & 4.50 & \textbf{84.1} & \textbf{96.8} \\
    \rowcolor{gray}
    \cellcolor{white} & \textbf{Castling-MViTv2-S} & 34.7 & 6.95 & \textbf{84.6} & \textbf{97.0} \\
  \hline
  \multirow{8}[2]{*}{$>$10G} & CaiT-S36~\cite{touvron2021going} & 68.0 & 13.90 & 83.3 & - \\
    & Swin-B~\cite{liu2021swin} & 88.0 & 15.40 & 83.5 & 96.5 \\
    & CSWin-B~\cite{dong2022cswin} & 78.0 & 15.00 & 84.2 & - \\
    & AutoFormer-B~\cite{chen2021autoformer} & 54.0 & 11.00 & 82.4 & 95.7 \\
    & MViTv2-B~\cite{li2022mvitv2} & 51.2 & 10.07 & 84.4 & - \\
    & DeiT-B~\cite{DeiT} & 86.3 & 17.56 & 83.4 & 96.5 \\
    \rowcolor{gray}
    \cellcolor{white} & \textbf{\hry{Castling-DeiT-B}} & 87.22 & 17.28 &  \textbf{84.2} & - \\
    \rowcolor{gray}
    \cellcolor{white} & \textbf{Castling-MViTv2-B} & 51.9 & 9.82 & \textbf{85.0} & \textbf{97.2} \\ 
  \hline
  \end{tabular}
  }
  \label{tab:imagenet}
  \vspace{-1.5em}
\end{table}

%% file: tables/coco_small.tex
\begin{table}[t]
  \centering
  \caption{Castling-ViT over SOTA baselines on COCO detection, where Castling-ViT$^{\dagger}$ means that the detection model adopts modified ESNet (use transformer blocks in the last stage; referred to ESNet-ViT) as backbone while Castling-ViT$^{\star}$ uses modified LCNet (use transformer blocks in the last two stage; referred to LCNet-ViT) as backbone. Also, ``-S/M/L'' and ``-320/416'' represent small/medium/large model variants and 320x320/416x416 input resolutions, respectively.}
  \vspace{-0.8em}
  \resizebox{\linewidth}{!}{
  \begin{tabular}{l|cc|ccc}
  \hline
  \textbf{Models} & \textbf{\tabincell{c}{Params\\(M)}} & \textbf{\tabincell{c}{MACs\\(G)}} & \textbf{mAP} & \textbf{AP50} & \textbf{AP75} \\
  \hline
  \hline
  MobileDet-DSP~\cite{xiong2020mobiledets} & 4.85 & 1.43 & 27.3 & - & - \\
  MobileDet-DSP-Fused~\cite{xiong2020mobiledets} & 9.15 & 3.22 & 29.1 & - & - \\
  EfficientDet-D0~\cite{tan2020efficientdet} & 3.90 & 2.54 & 34.6 & 53.0 & 37.1 \\
  \hline
  YOLOv5-N & 1.90 & 2.30 & 28.0 & 45.7 & - \\
  YOLOX-Nano~\cite{ge2021yolox} & 0.91 & 0.54 & 25.8 & - & - \\
  YOLOX-Tiny~\cite{ge2021yolox} & 5.06 & 3.23 & 32.8 & - & - \\
  \hline
  FBNetV5-AC-224x320~\cite{wu2021fbnetv5} & - & 0.71 & 25.0 & - & - \\
  FBNetV5-AR-224x320~\cite{wu2021fbnetv5} & - & 0.91 & 26.4 & - & - \\
  FBNetV5-A-224x320~\cite{wu2021fbnetv5} & - & 1.35 & 27.2 & - & - \\
  FBNetV5-AC-320x640~\cite{wu2021fbnetv5} & - & 1.37 & 28.9 & - & - \\
  FBNetV5-AR-320x640~\cite{wu2021fbnetv5} & - & 1.80 & 30.4 & - & - \\
  \hline
  \rowcolor{gray}
  \textbf{Castling-ViT-S-320$^{\dagger}$} & \textbf{3.25} & \textbf{0.62} & \textbf{28.1} & \textbf{42.3} & \textbf{29.2} \\
  \rowcolor{gray}
  \textbf{Castling-ViT-S-416$^{\dagger}$} & \textbf{3.25} & \textbf{1.06} & \textbf{31.3} & \textbf{46.7} & \textbf{32.5} \\
  \rowcolor{gray}
  \textbf{Castling-ViT-M-416$^{\dagger}$} & \textbf{6.01} & \textbf{2.00} & \textbf{34.0} & \textbf{49.5} & \textbf{35.9} \\
  \rowcolor{gray}
  \textbf{Castling-ViT-L-416$^{\dagger}$} & \textbf{9.32} & \textbf{3.03} & \textbf{35.0} & \textbf{50.7} & \textbf{37.2} \\
  \rowcolor{gray}
  \textbf{Castling-ViT-L-416$^{\star}$} & \textbf{13.10} & \textbf{5.31} & \textbf{37.3} & \textbf{53.4} & \textbf{39.6} \\
  \hline
  \end{tabular}%
  }
  \label{tab:coco_small}
  \vspace{-1.5em}
\end{table}

%% file: tables/ade20k.tex
\begin{table}[t]
  \centering
  \caption{Castling-ViT over baselines on the ADE20K segmentation dataset, serving as backbones in the Mask2former framework. Note that we report both total MACs and backbone MACs, i.e., ($\cdot$).}
  \vspace{-0.8em}
  \resizebox{\linewidth}{!}{
  \begin{tabular}{l|c|cc|ccc}
  \hline
  \textbf{Mask2former w/} & \multicolumn{1}{c|}{\textbf{\tabincell{c}{MAE\\Pretrain}}} & \textbf{\tabincell{c}{Params\\(M)}} & \multicolumn{1}{c|}{\textbf{\tabincell{c}{MACs\\(G)}}} & \textbf{mIoU} & \textbf{mAcc} & \textbf{pAcc} \\
  \hline
  \hline
  ViT-Base & N & 118 & 229 (182) & 34.54 & 46.36 & 75.84 \\
  \textbf{Castling-ViT-Base} & N & 118 & \textbf{195 (147)} & \textbf{34.67} & \textbf{46.47} & \textbf{76.20} \\
  \hline
  ViT-Base & Y & 118 & 229 (182) & 47.92 & 61.00  & 83.02 \\
  \textbf{Castling-ViT-Base} & Y & 118 & \textbf{195 (147)} & \textbf{48.44} & \textbf{61.82} & \textbf{83.29} \\
  \hline
  \end{tabular}%
}
\label{tab:ade20k}
\vspace{-0.6em}
\end{table}


%% file: tables/ablation_linear_attn.tex
\begin{table}[t]
  \centering
  \caption{Ablation studies of Castling-ViT with various kernel-based linear attentions, where MC refers to  mean-centering kernel function applied to $\mathbf{Q/K}$; Softmax also means the kernel function; ReLU-S and ReLU-E denote applying ReLU as kernels for decomposing softmax or exponential terms, respectively.
  All models are trained on COCO from scratch without pretraining.}
  \vspace{-0.8em}
  \resizebox{\linewidth}{!}{
  \begin{tabular}{l|l|cc|cc|cc}
  \hline
  \multicolumn{1}{l|}{\multirow{1}[4]{*}{\textbf{Backbone}}} & \multirow{1}[4]{*}{\textbf{Linear Attn}} & \multicolumn{2}{c|}{\textbf{S-320}} & \multicolumn{2}{c|}{\textbf{S-416}} & \multicolumn{2}{c}{\textbf{M-416}} \\
\cline{3-8}    &   & \textbf{mAP} & \textbf{AP50} & \textbf{mAP} & \textbf{AP50} & \textbf{mAP} & \textbf{AP50} \\
  \hline
  \hline
  \multicolumn{1}{l|}{LCNet} & N/A & 25.5 & 39.0 & 29.7 & 44.2 & 34.3 & 49.9 \\
  \multicolumn{1}{l|}{LCNet-ViT} & N/A & 25.9 & 39.6 & 29.8 & 44.4 & 34.9 & 50.4 \\
  \hline
  \multicolumn{1}{l|}{\multirow{6}[2]{*}{LCNet-ViT}} & MC~\cite{VITALITY} & 22.8 & 34.9 & 26.3 & 39.6 & 31.5 & 46.1 \\
    & Softmax~\cite{shen2021efficient} & 26.0 & 39.6 & 29.2 & 43.4 & 34.4 & 49.8 \\
    & Cosine~\cite{bolya2022hydra} & 24.4 & 37.4 & 27.2 & 41.0 & 32.9 & 48.0 \\
    & ReLU-S & 23.9 & 36.7 & - & - & 30.9 & 45.3 \\
    & ReLU-E~\cite{cai2022efficientvit} & 26.5 & 40.2 & 30.2 & 44.7 & 34.3 & 49.7 \\
    \rowcolor{gray}
    & \textbf{Lin.-Angular} & \textbf{27.0} & \textbf{40.6} & \textbf{30.4} & \textbf{45.5} & \textbf{35.5} & \textbf{51.1} \\
  \hline
  \end{tabular}%
}
\label{tab:ablation_linear_attn}
\vspace{-0.8em}
\end{table}

%% file: tables/ablation_s_320.tex
\begin{table}[t]
  \centering
  \caption{Ablation studies of Castling-ViT-S-320 w/ LCNet backbone, or LCNet-ViT backbone + linear-angular attention (denoted as Lin.) + DWConv (denoted as DW.) + auxiliary softmax-based attention (denoted as SparseAttn.).}
  \vspace{-0.8em}
  \resizebox{\linewidth}{!}{
  \begin{tabular}{l|cc|ccc}
  \hline
  \textbf{Castling-ViT-S-320 w/} & \multicolumn{1}{c}{\textbf{\tabincell{c}{Params\\(M)}}} & \multicolumn{1}{c|}{\textbf{\tabincell{c}{MACs\\(G)}}} & \multicolumn{1}{c}{\textbf{mAP}} & \multicolumn{1}{c}{\textbf{AP50}} & \multicolumn{1}{c}{\textbf{AP75}} \\
  \hline
  \hline
  LCNet & \multicolumn{1}{c}{1.10} & \multicolumn{1}{c|}{0.48} & \multicolumn{1}{c}{25.5} & \multicolumn{1}{c}{39.0} & \multicolumn{1}{c}{26.5} \\
  \hline
  LCNet-ViT & \multicolumn{1}{c}{2.14} & \multicolumn{1}{c|}{0.69} & \multicolumn{1}{c}{25.9} & \multicolumn{1}{c}{39.6} & \multicolumn{1}{c}{26.9} \\
  \textbf{+ Lin.} & \multicolumn{1}{c}{2.14} & \multicolumn{1}{c|}{0.62} & \multicolumn{1}{c}{26.2} & \multicolumn{1}{c}{39.7} & \multicolumn{1}{c}{27.3} \\
  \textbf{+ Lin.+DW.} & \multicolumn{1}{c}{2.14} & \multicolumn{1}{c|}{0.62} & \multicolumn{1}{c}{26.8} & \multicolumn{1}{c}{40.7} & \multicolumn{1}{c}{28.0} \\
  \textbf{+ Lin.+DW.+SparseAttn.} & \multicolumn{1}{c}{2.14} & \multicolumn{1}{c|}{0.62} & \multicolumn{1}{c}{27.0} & \multicolumn{1}{c}{40.6} & \multicolumn{1}{c}{28.0} \\
  \hline
  \end{tabular}%
}
\label{tab:ablation_s_320}
\vspace{-0.8em}
\end{table}

%% file: tables/ablation_m_416.tex
\begin{table}[t]
  \centering
  \caption{Ablation studies of Castling-ViT-M-416 w/ LCNet backbone, or LCNet-ViT backbone + Lin. + DW. + SparseAttn.}
  \vspace{-0.8em}
  \resizebox{\linewidth}{!}{
  \begin{tabular}{l|cc|ccc}
  \hline
  \textbf{Castling-ViT-M-416 w/} & \multicolumn{1}{c}{\textbf{\tabincell{c}{Params\\(M)}}} & \multicolumn{1}{c|}{\textbf{\tabincell{c}{MACs\\(G)}}} & \multicolumn{1}{c}{\textbf{mAP}} & \multicolumn{1}{c}{\textbf{AP50}} & \multicolumn{1}{c}{\textbf{AP75}} \\
  \hline
  \hline
  LCNet & \multicolumn{1}{c}{3.32} & \multicolumn{1}{c|}{2.15} & \multicolumn{1}{c}{34.3} & \multicolumn{1}{c}{49.9} & \multicolumn{1}{c}{36.5} \\
  \hline
  LCNet-ViT & \multicolumn{1}{c}{7.53} & \multicolumn{1}{c|}{3.51} & \multicolumn{1}{c}{34.9} & \multicolumn{1}{c}{50.4} & \multicolumn{1}{c}{37.0} \\
  \textbf{+ Lin.} & \multicolumn{1}{c}{7.53} & \multicolumn{1}{c|}{3.15} & \multicolumn{1}{c}{34.8} & \multicolumn{1}{c}{50.3} & \multicolumn{1}{c}{36.7} \\
  \textbf{+ Lin.+DW.} & \multicolumn{1}{c}{7.53} & \multicolumn{1}{c|}{3.15} & \multicolumn{1}{c}{35.1} & \multicolumn{1}{c}{50.9} & \multicolumn{1}{c}{36.9} \\
  \textbf{+ Lin.+DW.+SparseAttn.} & \multicolumn{1}{c}{7.53} & \multicolumn{1}{c|}{3.15} & \multicolumn{1}{c}{35.3} & \multicolumn{1}{c}{51.1} & \multicolumn{1}{c}{37.3} \\
  \hline
  \end{tabular}%
}
\label{tab:ablation_m_416}
\vspace{-1.em}
\end{table}

%% file: sections/5-Conclusion.tex
\vspace{-0.5em}
\section{Conclusion}
\label{sec:conclusion}
\vspace{-0.3em}


We present Castling-ViT that trains ViTs with both linear-angular and softmax-based quadratic attention but switches to only having the former during inference. Castling-ViT leverages angular kernels to measure the similarities between the queries and keys via spectral angles and highlights two enablers:
(1) a new linear-angular attention mechanism: we decompose angular kernels to linear terms and high-order residuals, and keep only the former for inference; and
(2) we approximate the high-order residuals using a depthwise convolution and an auxiliary masked softmax attention whose masks gradually become zeros during training without incurring inference overhead.
Extensive experiments consistently validate Castling-ViT's advantages.

\vspace{-0.5em}
\section*{Acknowledgment}
\vspace{-0.2em}
The work is supported in part by the National Science Foundation (NSF) RTML program (Award number: 1937592) and the CoCoSys, one of the seven centers in JUMP 2.0, a Semiconductor Research Corporation (SRC) program sponsored by DARPA.

%% file: sections/6-Supple.tex
\appendix

\section{More Literature Review}

\hry{
\textbf{Efficient ViTs.}
As previously mentioned in Sec. \ref{sec:related_works}, efficient attention can be roughly categorized into two groups: local attention~\cite{liu2021swin,arar2022learned, wang2020linformer,tu2022maxvit} or kernel-based linear attention~\cite{katharopoulos2020transformers,choromanski2021rethinking,xiong2021nystromformer,lu2021soft,bolya2022hydra,cai2022efficientvit,zhen2022cosformer,liu2022ecoformer,ali2021xcit}.
For local attention, Swin~\cite{liu2021swin} restricts the window size of self-attention, so that only neighboring tokens will perform similarity measurements each other instead of all tokens;
MaxViT~\cite{tu2022maxvit} also adopts block attention within windows but additionally takes dilated global attention into account for learning both local and global information;
QnA~\cite{arar2022learned} shares the attention queries among all tokens; Linformer~\cite{wang2020linformer} approximates the queries and keys with low-rank factorization to reduce their vector length.
For kernel-based linear attention,
XCiT~\cite{ali2021xcit} proposes a ``transposed'' version of self-attention that operates across feature channels rather than tokens, resulting in linear complexity with the number of tokens;
Linformer~\cite{wang2020linformer} explores a low-rank matrix to approximate the self-attention;
Reformer~\cite{kitaev2020reformer} replaces self-attention by one that uses locality-sensitive hashing, changing its complexity from $\mathcal{O}(n^2)$ to $\mathcal{O}(n\log(n))$ where $n$ denotes the number of tokens;
Longformer~\cite{beltagy2020longformer} combines a windowed local-context self-attention and a task-motivated global attention that encodes inductive bias about that task;
Nystromformer~\cite{xiong2021nystromformer} adapts the Nystrom method
to approximate standard self-attention with $\mathcal{O}(n)$ complexity; 
Scatterbrain~\cite{chen2021scatterbrain} unifies both low-rank approximation and sparse attention to improve accuracy-efficiency tradeoffs.
Different from all above works, we explore from a new perspective by taking spectral angles into consideration when measuring the similarity among tokens, resulting in linear-angular attention with sparse training techniques that can achieve comparable or even better performance than softmax-based attention.
}

\textbf{ViTs for Downstream Tasks.}
Apart from image classification tasks, ViTs have also been leveraged to serve as backbones for downstream tasks, such as object detection~\cite{carion2020end,zhu2020deformable,PVT_ICCV,li2022mvitv2,zhang2021vit} and semantic segmentation~\cite{cheng2021per,qin2022pyramid,cheng2022masked}.
For example, DETR~\cite{carion2020end} directly detects and predicts objects by combining a common CNN with a transformer architecture;
Maskformer~\cite{cheng2021per} proposes to use a simple mask classification model to predict a set of binary masks, each associated with a single global class label prediction.
One big difference is that ViTs can beat CNNs on classification tasks that have a lower image resolution while are still less efficient than lightweight CNNs on downstream tasks that heavily rely on multi-scale resolution features.
Therefore, there have been various debates on designing powerful ViT backbones: (1) \textit{plain ViTs (e.g., ViTDet~\cite{li2022exploring}) or hierarchical ViTs (e.g., MViTv2~\cite{li2022mvitv2}, or Swin~\cite{liu2021swin})?} Plain ViTs win in terms of simplicity but could be hard to scale down to lower resolution and computation regimes; Hierarchical ViTs seamlessly match with feature pyramid networks (FPNs) for extracting multi-scale feature maps but have more design factors to be considered or searched over.
(2) \textit{pure ViTs or hybrid CNN-ViTs?} Pure ViTs are compatible with self-supervised masked autoencoder (MAE) pretraining~\cite{he2022masked}; Hybrid CNN-ViTs can suffer from the information leakage problem~\cite{gao2022convmae} when adopting MAE pretraining, while being more efficient especially for feature extractions in early layers.
Our proposed method does not fall into the aforementioned debates. Instead, it is compatible with all ViT variants relying on the softmax-based attention.

\section{More Results and Clarification}
\label{sec:supply}

\textbf{Improvement from Our Training Recipe.} Recall that in Sec. \ref{sec:exps}, we conduct experiments on three classical computer vision tasks. For object detection and semantic segmentation, we follow the  baseline's training recipe for a fair and direct  comparison. For the image classification, our training recipe has a minor difference due to the increased batch size and training epochs with more GPU nodes. As such, we further provide the detailed improvement breakdown here. Specifically, our adopted training recipe leads to $\uparrow$0.2\% $\sim$ $\uparrow$1.6\% top-1 accuracy improvements and our Caslting-ViT further reduces up to $\downarrow$40\% MACs and increases $\uparrow$0.1\% $\sim$ $\uparrow$1.2\% top-1 accuracy simultaneously.



\begin{table*}[t]
    \centering
    \caption{Throughputs/memory measurements on a V100 for image classification, under various input resolutions denoted as $r \times r$.}
    \resizebox{0.75\linewidth}{!}{
      \begin{tabular}{c|ccc|ccc}
      \hline
      \multirow{2}[4]{*}{\textbf{Model}} & \multicolumn{3}{c|}{\textbf{Throughputs (Images/s)}} & \multicolumn{3}{c}{\textbf{GPU Peak Memory (MB)}} \\
      \cline{2-7}    & \textbf{$r$ = 512} & \textbf{$r$ = 1024} & \multicolumn{1}{c|}{\textbf{$r$ = 1536}} & \textbf{$r$ = 512} & \textbf{$r$ = 1024} & \multicolumn{1}{c}{\textbf{$r$ = 1536}} \\
      \hline
      \hline
      DeiT-Base & 40  & 6  & OOM  & 1220  &  12369 & OOM \\
      Castling-DeiT-Base & 48 ($\uparrow$20\%) & 8 ($\uparrow$33\%)  &  6 & 998 ($\downarrow$18\%)  & 4863 ($\downarrow$61\%) & 15478\\
      \hline
      MViTv2-Base & 43  & 5  & OOM & 1762  & 14686 & OOM \\
      Castling-MViTv2-Base & 50 ($\uparrow$16\%) & 10 ($\uparrow$100\%)  & 4  & 1483 ($\downarrow$16\%)  & 7028 ($\downarrow$52\%) & 16028 \\
      \hline
      \end{tabular}%
    }
    \label{tab:throughputs}
\end{table*}

\hry{
\textbf{Ablation Studies on Image Classification.}
Our ablation studies are mostly done on the detection task as shown in Sec. \ref{sec:ablation_castling_vit} because of its less training time as compared to training ImageNet.
Note that for these ablation studies, we do not adopt pretraining on ImageNet as specified in Sec. \ref{sec:ablation_castling_vit}.
After finishing the trial-and-error and when it comes to comparing with SOTA works, we then pretrain final models with the training recipe the same as LeViT~\cite{graham2021levit}, resulting in final results in Tab. \ref{tab:coco_small}.
According to our experiments, training ImageNet takes nearly one week, while training COCO without pretraining on ImageNet takes only one day.
In fact, ablation results on the classification task are consistent.
To deliver more comprehensive ablation studies, we train Castling-LeViT-256 on ImageNet afterwards and find that:
(1) + Lin.: 81.5\%; (2) + Lin. \& DWConv: 82.4\%; (3) + Lin. \& DWConv \& SparseAttn: 82.6\%, those results are consistent with our observation on detection experiments.
}

\hry{
\textbf{Conjecture of Why Linear-Angular Attention Sometimes Beats the Original Self-Attention.}
To better understand why the result of our Castling-ViT is even better than softmax-based ViTs. We summarize three differences between our method and previous linear attentions:
(1) In addition to linear attention, we also take DWConv and sparse softmax-based attention into the training process;
(2)
We use a SGD optimizer instead of Adam, which is not common for training ViTs. Although Adam optimizer leads to faster convergence, we find that SGD optimizer helps to deliver better results if being trained sufficiently converged, e.g., we train 1000 epochs on ImageNet;
(3) After revisiting the attention design, we remove token/feature pooling and adopt post-$\mathbf{Q}$ pooling and residual connections~\cite{li2022mvitv2} in our attention blocks.
All above three differences contribute to the the final accuracy apart from the improvement of using linear-angular attention. We also show the breakdown analysis for each of these three points, see Sec. \ref{sec:ablation_castling_vit}, Sec. \ref{sec:supply}, and Sec. \ref{sec:preliminary} for detailed analysis, respectively.
}

\begin{table}[t]
    \centering
    \caption{Ablation study on the patch size (measured on a V100).}
    \resizebox{\linewidth}{!}{
      \begin{tabular}{A|IIB}
      \hline
      \multirow{2}[4]{*}{\textbf{Models}} & \multicolumn{3}{c}{\textbf{Throughputs (Images/s) under $p$ patch sizes}} \\
      \cline{2-4}   & \textbf{$p$ = 8} & \textbf{$p$ = 4} & \textbf{$p$ = 2} \\
      \hline
      \hline
      DeiT-Tiny  & 398  & 40  &  3 \\
      Castling-DeiT-Tiny & 410 ($\uparrow$1.0$\times$) & 103 ($\uparrow$2.6$\times$) & 20 ($\uparrow$6.7$\times$) \\
      \hline
      DeiT-Base & 60  & 8  &  OOM \\
      Castling-DeiT-Base & 64 ($\uparrow$1.1$\times$)  & 15 ($\uparrow$1.9$\times$) &  4\\
      \hline
      \end{tabular}%
    }
    \label{tab:patch}
    \vspace{-0.5em}
\end{table}

\hry{
\textbf{Actual Latency, Throughputs, and Memory Measurements.}
Our final models are dense and thus well compatible with GPUs. We measure and report the latency ($\downarrow$55\%), throughputs ($\uparrow$16 $\sim$ $\uparrow$100\%), and GPU memory ($\downarrow$16 $\sim$ $\downarrow$61\%) for both classification and detection tasks, as shown in Tab. \ref{tab:throughputs}/\ref{tab:detection}.
For throughputs, we measure both our Castling-ViT and baselines under their maximum allowed batch sizes (bs), i.e., bs=16/2/1, for different input resolutions $r$=512/1024/1536 in a fair and consistent V100 environment.
Note that when the input resolution $r$=224, our models cannot beat the baseline in terms of throughputs because of (1) the newly added DWConv; (2) the removal of token/feature pooling.
However, in terms of accuracy-efficiency tradeoffs, our Castling-ViT consistently beats all baselines as shown in Sec. \ref{sec:exps}.
For memory, we record the peak memory per image.
For latency, 
we benchmark with SOTA CNN-based detectors. Our model achieves 37.3mAP at 3.9ms latency on a V100, while YOLOv5-S only achieves 36.7mAP at 8.7ms latency).
Moreover, Castling-ViT wins more throughputs (up to $\uparrow$6.7$\times$) for smaller patch sizes and/or larger input resolutions, as shown in Tab. \ref{tab:patch} and \ref{tab:throughputs}.
Note that we record CUDA latency following the literature~\cite{DeiT,liu2021swin}. All reported results are averaged among three runs.
}

\begin{table}[t]
    \centering
    \caption{Latency measurements on a V100 for object detection.}
    \resizebox{\linewidth}{!}{
      \begin{tabular}{c|cc|cc}
      \hline
      \textbf{Models} & \textbf{\tabincell{c}{Params\\(M)}} & \textbf{\tabincell{c}{MACs\\(G)}} & \textbf{mAP} & \textbf{\tabincell{c}{Latency\\(ms)}}  \\
      \hline
      \hline
      YOLOv5-S & 7.3 & 17.1 & 36.7  & 8.7~\cite{ge2021yolox} \\
      RetinaNet+PVT-Tiny~\cite{wang2021pyramid} & 23.0 & 221 & 36.7 & - \\
      Castling-ViT-L-416 & 13.1 & 5.3 & 37.3 & 3.9 ($\downarrow$55.2\%) \\
      \hline
      \end{tabular}%
    }
    \label{tab:detection}
\end{table}

\hry{
\textbf{Compare with ViT-based Baselines on Detection.}
We benchmark with SOTA CNN-based detectors under 6G MACs in Sec. \ref{sec:exps} because that ViT-based detectors are too expensive. For example, our Castling-ViT achieves 37.3mAP at 5.3G MACs, while RetinaNet+PVT-Tiny only achieves 36.7mAP at even 221G MACs~\cite{wang2021pyramid}, as shown in \ref{tab:detection}.
}

\hry{
\textbf{Advantages of Angular Kernels?}
Angular kernels take into account extra spectral characteristics and enjoy good properties, e.g., positive semi-definite function $\rightarrow$ inner product in a high-dimensional and rich feature space, as analyzed in Sec. \ref{sec:castling-vit}.
It also achieves comparable accuracy with vanilla attention as validated by Sec. \ref{sec:exps}.
}

\hry{
\textbf{Large-Scale Ablation Studies on Attention Design.}
We use small ViTs for idea validation in Tab. \ref{tab:ablation_linear_attn} and the conclusion generalizes to larger ones.
Here we add another ablation study on a larger model LeViT-384 as shown in Tab. \ref{tab:levit384}, from which we see that the attention design insights consistently generalize from small models to larger models, further validating our design insights.
}

\hry{
\textbf{Why More Parameters Than Others in Low MACs?}
ViTs tend to have more parameters than CNNs under small MACs, e.g., LeViT~\cite{graham2021levit} and Efficient-ViT~\cite{cai2022efficientvit}.
For the LeViT, it features more layers with gradually downsampled input resolutions.
For example, LeViT-256 requires 18.9M parameters at only 1.1G MACs, LeViT-384 requires 39.1M parameters at 2.4G MACs.
Since we adopt LeViT-like structure to construct our Castling-ViT on image classification tasks, the parameter looks higher than other else baselines.
Also, as indicated in Sec. \ref{sec:preliminary}, Castling-LeViT uses merely post-$\mathbf{Q}$ pooling, causing slightly higher hidden dimensions for $\mathbf{Q}$/$\mathbf{K}$ than LeViT.
In this work, we focus more on the FLOPs/latency instead of parameters since storage is not a major concern in modern hardware~\cite{schaller1997moore}.
}

\hry{
\textbf{Will Auxiliary Attention and DWConv Work for Existing Linear Attentions?}
Yes, we train a DeiT-Tiny (w/o distill.; Acc.: 72.2\%) w/ linear attention~\cite{VITALITY} for 300 epochs and observe that:
(1) + Lin.: 68.3\%; (2) + Lin. \& DWConv: 71.7\%; (3) + Lin. \& SparseAttn: 70.2\%; (4) + Lin. \& DWConv \& SparseAttn: 72.4\%.
}

\begin{table}[t]
    \centering
    \caption{Additional ablation studies on attention designs using LeViT-384 on ImageNet.}
    \resizebox{\linewidth}{!}{
      \begin{tabular}{cc|cc|c|c|c}
      \hline
      \multicolumn{4}{c|}{\textbf{Pooling}} & \multirow{2}[1]{*}{\textbf{Residual} $\mathbf{Q}$} & \multirow{2}[1]{*}{\textbf{MACs (G)}} & \multirow{2}[0]{*}{\textbf{Top-1 Accuracy (\%)}} \\
    \cline{1-4}  \textbf{Token} & \textbf{Feat.} & \textbf{Pre-$\mathbf{Q}$} & \textbf{Post-$\mathbf{Q}$} &   &   &  \\
      \hline
      \hline
      \Checkmark &   & \Checkmark &   &   & 2.50 & 82.63  \\
        & \Checkmark & \Checkmark &   &   & 2.36 & 82.55 \\
        &   & \Checkmark &   &   & 2.61 & 82.65  \\
      \hline
      \Checkmark &   &   & \Checkmark & \Checkmark & 2.83 & 82.19 \\
        & \Checkmark &   & \Checkmark & \Checkmark & 2.80 & 81.86 \\
        &   &   & \Checkmark & \Checkmark & 2.96 & 83.65 \\
      \hline
      \end{tabular}%
    }
    \label{tab:levit384}
\end{table}

\hry{
\textbf{Clarify ReLU-S vs. ReLU-E in Tab. \ref{tab:ablation_linear_attn}.}
During approximation, i.e.,
$\text{Sim}(\mathbf{Q}, \mathbf{K}) \approx \phi(\mathbf{Q}) \phi(\mathbf{K})^T$,
both of them use ReLU as $\phi(\cdot)$, but ReLU-S takes the whole Softmax as $\text{Sim}(\cdot)$, while ReLU-E takes the $\text{Exp}(\cdot)$ as $\text{Sim}(\cdot)$, e.g., Efficient-ViT, resulting in additional divisions.
}